\documentclass{article}

\usepackage[english]{babel}
\usepackage[T1]{fontenc}

\usepackage[letterpaper,top=2cm,bottom=2cm,left=3cm,right=3cm,marginparwidth=1.75cm]{geometry}

\usepackage{amsmath}
\usepackage{amssymb}
\usepackage{graphicx}
\usepackage[dvipsnames]{xcolor}
\usepackage{fullpage}
\usepackage{csquotes}
\usepackage{booktabs}
\usepackage{bm}
\usepackage{enumitem}
\usepackage{calc}
\usepackage{multirow}
\usepackage{makecell}
\usepackage{caption}
\usepackage{subcaption}
\usepackage{siunitx}

\usepackage{enumitem,amssymb}
\usepackage{pifont}
\usepackage{authblk}

\usepackage{hyperref}
\hypersetup{
    colorlinks,
    linkcolor={red!70!black},
    citecolor={blue!80!black},
    urlcolor={cyan!50!black}
}

\DeclareMathOperator{\prob}{\text{Pr}}
\DeclareMathOperator*{\argmin}{arg\,min}

\title{Probabilistic load forecasting with Reservoir Computing}
\date{}
\author[1]{Michele Guerra\thanks{Corresponding Author: michele.guerra@uit.no}}
\author[2]{Simone Scardapane}
\author[1,3]{Filippo Maria Bianchi}
\affil[1]{UiT the Arctic University of Norway}
\affil[2]{Sapienza University of Rome}
\affil[3]{NORCE Norwegian Research Centre}

\begin{document}

\maketitle


\subsection*{Abstract} 
Some applications of deep learning require not only to provide accurate results but also to quantify the amount of confidence in their prediction. The management of an electric power grid is one of these cases: to avoid risky scenarios, decision-makers need both \emph{precise} and \emph{reliable} forecasts of, for example, power loads. For this reason, point forecasts are not enough hence it is necessary to adopt methods that provide an \emph{uncertainty quantification}.

This work focuses on reservoir computing as the core time series forecasting method, due to its computational efficiency and effectiveness in predicting time series.
While the RC literature mostly focused on point forecasting, this work explores the compatibility of some popular uncertainty quantification methods with the reservoir setting. Both Bayesian and deterministic approaches to uncertainty assessment are evaluated and compared in terms of their prediction accuracy, computational resource efficiency and reliability of the estimated uncertainty, based on a set of carefully chosen performance metrics.



\section{Introduction}

Models for time-series forecasting are crucial tools in many fields, such as climate modelling, finance, meteorology and energy analytics~\cite{lim2021time}.
For example, an accurate prediction of the energy demand is fundamental to managing a power grid and preventing disruptions in the energy supply, which might cause serious and costly consequences~\cite{eikeland2021detecting}. 
Especially with a larger penetration of intermittent energy resources~\cite{antonanzas2016review, wang2011review},
load forecasting plays a key role in the efficient planning of electricity supplies~\cite{liu2017electricity}.

Other than producing a forecast, it is important to quantify the uncertainty in the predictions, since that affects how reliable the forecast is for critical decision-making~\cite{gal2016uncertainty}.
For example, if the model is fed with an input that lies far from the distribution of training data, it is desirable that it provides a high uncertainty about its prediction, to alert the user about a possibly unreliable prediction.
In this regard, a method that computes point estimates is not enough. On the other hand, probabilistic forecasting methods provide the whole distribution of the forecast or, at least, some confidence intervals~\cite{jensen2022ensemble, du2023probabilistic}.

Recurrent Neural Networks (RNN) are one of the most effective models for processing data with causal dependencies such as time series, but they come with some drawbacks.
Those include the difficulty of modelling long-term dependencies when performing gradient descent through time and the long training times due to their sequential structure that prevents parallelisation on modern hardware~\cite{bianchi2017recurrent}. 
Reservoir computing overcomes these problems by leaving the recurrent part of the model untrained~\cite{lukovsevivcius2009reservoir}, which allows bypassing the propagation of the gradient through time, greatly reducing the training time. Remarkably, reservoir computing models perform as well as deep learning methods on several tasks~\cite{gauthier2021next}.
This work uses an Echo State Network (ESN) as a core forecasting model for power demand time series. An ESN is a reservoir computing (RC) model particularly suited for systems that display complex and chaotic behaviour~\cite{shahi2022prediction}. As explained in more detail later, the ESN embeds the input in a high dimensional vector state which can then be fitted to the desired output. 

So far, most of the existing forecasting approaches based on ESNs have focused on computing point estimates, limiting their use in real-world applications that require estimating confidence intervals for the predictions.
To fill this gap, we couple the ESN with popular methods of uncertainty quantification and we adapt them to the reservoir computing paradigm, especially by studying how to deal with the high-dimensional embeddings generated by the reservoir.
We provide an extensive comparison in terms of forecasting accuracy, computational costs and quality of the produced uncertainty using multiple metrics.
In particular, we carried out our experiments on two real-world datasets, representing the electric load of two power grids.
Our findings suggest that popular Bayesian methods, such as Markov chain Monte Carlo and variational inference, suffer considerably from the high dimensionality of the reservoir embeddings, while quantile regression proves to be accurate and lightweight.
To the best of our knowledge, a comprehensive study that thoroughly evaluates uncertainty quantification in reservoir computing does not exist and it could provide valuable insights to a practitioner looking for models with good trade-offs between performance, training time and uncertainty estimation.

The paper is organised as follows. Section~\ref{sec:related} presents a summary of the most recent literature on the subject, then a background section introduces reservoir computing~\ref{sec:rc} and the problem of quantifying uncertainties~\ref{sec:uncertainty}. All the methods we are comparing are presented in Section~\ref{sec:methods}, finally in Sections~\ref{sec:experiments} and~\ref{sec:results} we describe the experiments and provide an evaluation of the results.

\section{Related works}
\label{sec:related}

In the last few decades, machine learning methods earned increasing popularity for time series forecasting.
In particular, RNNs became one of the most popular options when working with data characterised by time dependencies~\cite{chandra2021}. Gated RNNs, such as LSTM~\cite{hochreiter1997long}, were introduced to address some limitations of vanilla RNNs~\cite{elman1988learning}. The reader can find comparisons of machine learning methods for time series forecasting in~\cite{chandra2021, lim2021time}. Reservoir computing, introduced in~\cite{jaeger2001echo}, proves to perform better than other machine learning approaches on time series stemming from complex dynamical systems~\cite{shahi2022prediction}.

The deep learning models mentioned so far were originally designed to perform point forecasting, which is the prediction of the most likely value, given a particular choice of the loss function.
The simplest approach to model uncertainty is to train an ensemble of RNNs and build confidence intervals from the statistics of the predictions in the ensemble~\cite{wang2022novel}. This approach makes strong assumptions about the underlying data distribution and often generates overconfident intervals if the models in the ensemble are not different enough~\cite{konstantinou2023day}.
A more elegant approach to model the uncertainty is quantile regression, which was first introduced in~\cite{koenker1978regression} and later further developed in~\cite{gasthaus2019probabilistic}. Quantile regression predicts pre-defined quantiles of the output distribution and can be easily implemented in a neural network trained by minimising a particular loss function~\cite{wang2019probabilisticlstmpinball}. 
Another approach, called conformal prediction, generates prediction intervals from the statistics obtained by a trained model on a calibration set. Originally designed for independent data~\cite{angelopoulos2021gentle}, conformal prediction has been recently extended to handle time series data~\cite{stankeviciute2021conformal} and paired with RNNs~\cite{xu2020conformal_dynamic, jensen2022ensemble}.

The interest in probabilistic forecasting with deep learning has recently increased, as evidenced by the presence of open-source libraries such as GluonTS~\cite{gluonts_jmlr} and PyTorch Forecasting\footnote{Source: \url{https://pytorch-forecasting.readthedocs.io/en/stable/}} that provide several models out of the box.
One of the deep learning models for probabilistic forecasting that gained popularity in recent years is DeepAR~\cite{salinas2020deepar}. DeepAR uses an LSTM that produces an embedding of the input time series and is paired with a probabilistic model chosen by the user, which forecasts future values in the form of Monte Carlo samples. A similar approach is followed by DeepTCN~\cite{chen2020probabilistic}, which uses temporal convolutional networks and outputs the parameters of a chosen probability distribution from which one can sample the forecast. In~\cite{qiu2020multivariate}, the authors use a Bayesian model trained with Markov chain Monte Carlo and a sparsity-inducing peak-and-slab prior over regression parameters. 
A recent overview of probabilistic forecasting based on deep learning models can be found in the related works section of~\cite{eikeland2022probabilistic}. 

Despite the prolific research on probabilistic forecasting with deep learning, only a few efforts have been devoted to implementing probabilistic forecasting with RC approaches. 
ESGP~\cite{chatzis2011echo} combines ESN with Gaussian processes, in~\cite{li2012chaotic} Bayesian regression with Gaussian and Laplace priors is used to train the ESN, while \cite{chatzis2015sparse} employs stochastic variational inference. Another work that goes in this direction is~\cite{mcdermott2019deep}, where the authors introduced a Bayesian deep ESN~\cite{jaeger2007discovering} and used MCMC to account for the prediction's uncertainty. To try to make the model lighter, they also make use of stochastic search variable selection priors. They also compare their proposed framework against ensemble forecast methods.
These four works consider a single Bayesian approach to model the uncertainty in the prediction of an ESN.
However, a systematic analysis and comparison between popular Bayesian and non-Bayesian uncertainty quantification techniques within an RC framework are still missing. The aim of the present work is to fill this gap.

\section{Background}
\label{sec:background}

\subsection{Reservoir computing}
\label{sec:rc}

The most popular RC model in machine learning is the ESN; the terms RC and ESN are often used interchangeably~\cite{jaeger2001echo}.
An RC model is composed of three elements: the input layer and the recurrent layer called \emph{reservoir}, which are randomly initialised and remain untrained, and a linear output layer, which is the only part that is trained, usually by linear regression.
Denoting with $x_t\in\mathbb{R}^K$, $y_t\in\mathbb{R}^L$ and $s_t\in\mathbb{R}^N$ respectively the input, output and state of the reservoir at time $t$, the ESN dynamics is governed by the following equations:
\begin{align}
    s_{t+1}&=f(W_\text{in}x_{t+1}+Ws_t) \\
    y_t&=g(R s_t),
\end{align}
where $W_\text{in}\in\mathbb{R}^{N\times K}$ and $W\in\mathbb{R}^{N\times N}$ are fixed and $R\in\mathbb{R}^{L\times N}$ is the only trainable part; the two functions $f$ and $g$ introduce non-linearities. 
In general, an ESN could also include feedback from the output in the equation that updates the reservoir state, but we are not considering it.

In order for the ESN to produce meaningful embeddings, the random matrix $W$ must be initialised ensuring that the echo state property is satisfied~\cite{gallicchio2011architectural, yildiz2012re}.
Loosely speaking, this requires that the state of the reservoir $s_t$ asymptotically should not depend on its initial state, but only on the input sequence.

A widely used rule of thumb to initialise the reservoir is to set its spectral radius to less than one.
However, being the ESN a non-autonomous system, it has been proved that this criterion is neither necessary nor sufficient to ensure the echo state property and it could also be far from the optimal choice~\cite{yildiz2012re}.
For this reason, more advanced methods have been proposed to encourage the reservoir to generate rich dynamics of the input~\cite{bianchi2016investigating, bianchi2017multiplex}. Nevertheless, it has been proven that limiting the spectral radius to be less than one is ``sufficient in practice'' since a reservoir built in this way satisfies the echo state property with high probability~\cite{zhang2011nonlinear}. 
Therefore, even if potentially sub-optimal, in this work we use the aforementioned criterion for the sake of simplicity and practicality.

Finally, while the vanilla ESN implements $g(s_t; R)$ as a linear readout, more recent works showed that a non-linear readout implemented by a multi-layer perceptron (MLP) usually achieves better performances~\cite{bianchi2018bidirectional}. In order to train $g$ to accurately map each state $s_t$ into an output $y_t=g(s_t; R)$, where $R$ represents the model's trainable parameters, a training dataset must be constructed. The input time-series $\{x_t\}_{t=1}^T$, with $x_t\in\mathbb{R}$, is used by the reservoir $f$ to update its inner state $S=\{s_t\}_{t=1}^T$; hence, the training dataset is defined as $\{(s_t,y_t)\}_{t=1}^T$, where $y_t=x_{t+h}$, with $h$ the forecast horizon.
As it is, $g$ is a deterministic function and deterministic models return only one data point for each input, so they can't possibly produce an estimate for the uncertainty.

\subsection{Bayesian approach} 
\label{sec:uncertainty}

In the present work, we focus on the readout $g(s_t; R)$ and how we can change it so as to provide uncertainties.
One approach is to modify the above setting so that the output of the model provides $y$ together with information on its uncertainty. This can be done, for example, with quantile regression, which will be introduced in Section~\ref{sec:qr}. Alternatively, one can resort to Bayesian Neural Networks (BNN), which naturally embody information about uncertainty by generating entire distributions over the output domain.

In a BNN, $R$ is described by a distribution that represents what are the values of $R$ that most likely generate data.
In this probabilistic setting, we are interested in searching for the posterior distribution $p(R|S,Y)$, where $S$ and $Y$ are random variables representing respectively the reservoir state and the output, which we could then use to sample an output $y^*$ for a previously unseen state $s^*$ from
\begin{equation}
    p(y^*|s^*,S,Y)=\int_\Omega p(y^*|s^*,R)p(R|S,Y) \,dR,
\end{equation}
where $\Omega$ is the domain of $R$ and $p(y|s,R)$ is the likelihood, and it is given by the model itself.\footnote{Notice that we use the same symbol, i.e.~$R$, to indicate both the random variable and a possible value that it might assume.}

Bayes' theorem gives the explicit form of the posterior as
\begin{equation}
    \label{eq:bayes}
    p(R|S,Y)=\frac{p(Y|S,R)p(R)}{p(Y|S)},
\end{equation}
where at the numerator we have the likelihood and the prior $p(R)$, which are known and selected by us.
A standard option for the likelihood that we use in our work is a Gaussian with variance $\Sigma$
\begin{equation}
    \label{eq:likelihood}
    Y \sim p(y|s,R)=\mathcal{N}(g(s;R),\Sigma).
\end{equation}
As priors for the parameters $R$,\footnote{Given our choice of the likelihood in~\eqref{eq:likelihood}, $\Sigma$ is also a parameter of the model. For ease of notation, we are not specifying it every time, and we only refer to $R$.} it is common to use a Gaussian or uniform distribution, according to the standard approach to use semi-informative priors~\cite{gelman2017prior, sarma2020prior}. A more specialised choice for $p(R)$ is the horseshoe distribution, discussed in Section~\ref{sec:mcmc}. 

The problem about~\eqref{eq:bayes} lies in the normalisation constant at the denominator
\begin{equation}
    \label{eq:normalisation}
    p(Y|S)=\int_\Omega p(Y|S,R)p(R)\,dR,
\end{equation}
which is often intractable and cannot be computed analytically.
This issue can be overcome by approximating the posterior with variational inference (see Section~\ref{sec:vi}) or by using numerical routines, like Markov chain Monte Carlo (see Section~\ref{sec:mcmc}).
Another way to create a BNN and avoid computing \eqref{eq:normalisation} is by transforming a deterministic model into a probabilistic one using dropout (see Section~\ref{sec:dropout}).
This latter approach sacrifices the fine control over the priors, which naturally allows the embedding of previous knowledge and acts as regularisation when dealing with small datasets~\cite{gelman2017prior}.

\section{Methods}
\label{sec:methods}

\begin{figure}
     \centering
     \includegraphics[width=\textwidth]{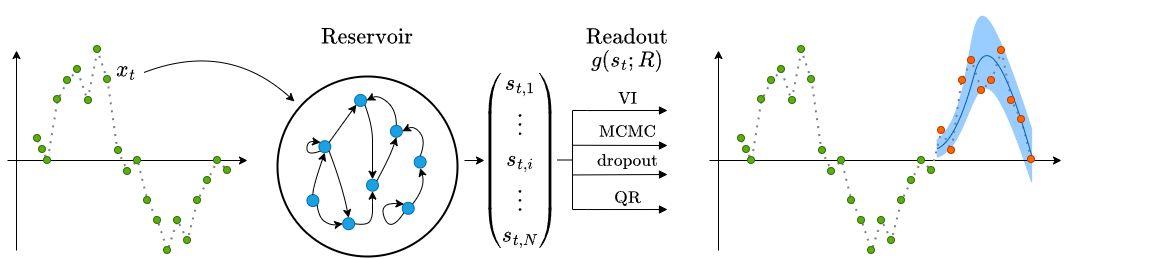}
     \caption{Overview of the used framework. On the left, in green, is the input time series used for training. Each point $x_t$ is fed into the reservoir and used to update its internal state $s_t$. The series of states is then used to train the readout $g(s_t;R)$ using one of the represented methods. On the right, we have the outcome, with ground truth values in orange, and a prediction represented by a solid blue line, with its uncertainty pictured as a shaded band around it.}
     \label{fig:framework}
 \end{figure}

What follows is a description of the methods to produce an uncertainty estimate that we compare in this work.
We start with the three approaches for training a BNN, then we present QR (Section~\ref{sec:qr}), and we conclude by discussing the problem of calibration in Section~\ref{sec:calibration}. Refer to Figure~\ref{fig:framework} for a schematic view of the framework.

\subsection{Variational inference}
\label{sec:vi}

Variational Inference (VI) offers a way to tackle the intractability of~\eqref{eq:bayes}. 
The idea is to substitute the true posterior $p(R|S,Y)$ with an approximation $q_\theta(R)$, called \emph{variational distribution}, which depends on some parameters $\theta$. In order to identify among all the possible $q_\theta(R)$ the one that best approximates the posterior, one has to minimise the Kullback–Leibler divergence between the true and approximate posterior
\begin{equation}
    \text{KL}(q_\theta(R)\,||\,p(R|S,Y)):=\int_\Omega q_\theta(R)\log\frac{q_\theta(R)}{p(R|S,Y)}\,dR.
\end{equation}
One can prove that
\begin{equation}
\label{eq:svi-optim}
    \text{KL}(q_\theta(R)\,||\,p(R|S,Y))\leq \text{KL}(q_\theta(R)\,||\,p(R)) - \mathcal{L}_\text{LL},
\end{equation}
where $\mathcal{L}_\text{LL}:=\int q_\theta(R)\log p(Y|S,R)\,dR$ is the expected log-likelihood. Hence an equivalent approach to minimising the KL on the left is to minimise the right-hand side (which does not depend anymore on $p(R|S,Y)$), i.e.~maximising the Evidence Lower Bound (ELBO), defined as
\begin{equation}
    \mathcal{L}_{\text{VI}}(\theta):=\mathcal{L}_\text{LL}-\text{KL}(q_\theta(R)||p(R)),
\end{equation}
where, by maximising the first term, we are pushing $q_\theta(R)$ to better approximate the data distribution, while the second term keeps it close to the prior $p(R)$. VI, in other words, translates the task of solving~\eqref{eq:bayes} into the optimisation problem~\eqref{eq:svi-optim}, which requires computing derivatives rather than integrals.

As variational distribution $q_\theta(R)$, we use a multivariate Gaussian distribution on the whole parameter space, constructed using the Cholesky decomposition, where we don’t assume any independence between parameters.

In our case, the parameters $R$ are mainly stemming from the MLP $g(s;R)$ used as a readout of the ESN, so their number $N$ increases with the number of units within each layer and with the number of layers. For example, if we consider a linear layer $500\times 100$, the dimension of $R$ for that layer alone will be $N=\num{50000}$. A multivariate Gaussian as variational distribution would then be defined by a covariance matrix of size $N^2$, which reserves a lot of computing memory, leading easily to its saturation.

To avoid this issue we adopted a sparsity-inducing distribution, like the \emph{low-rank} multi-variate Gaussian, as variational distribution $q_\theta(R)$. If we assume that $R\in\mathbb{R}^N$ depends on a smaller number of latent parameters $\phi\in\mathbb{R}^r$, we have $R=C\phi+\epsilon$, where $C$ is a $N\times r$ matrix with $r<N$, $\phi$ is a $r$-dimensional random variable from a Gaussian distribution with 0 mean and covariance $\mathbb{I}_r$, and $\epsilon$ is a $N$-dimensional vector of independently distributed error terms with zero mean and finite variance $\text{var}(\epsilon)=\Psi$. Then one can prove that
\begin{equation}
    \text{cov}(R)=CC^T+\Psi,
\end{equation}
so the covariance matrix can be expressed as the sum of a low-rank matrix $CC^T$ of rank $r$ and a diagonal matrix $\Psi$. A common choice is $r=\sqrt{N}$. Notice that when $C=0$ the covariance matrix becomes diagonal, so the mean-field approximation, where all parameters $R$ are assumed independent, is a special case.

\subsection{Markov chain Monte Carlo}
\label{sec:mcmc}

Another way to tackle the posterior in~\eqref{eq:bayes} is to use Markov Chain Monte Carlo (MCMC), which allows sampling from the posterior without the need to compute the integral in~\eqref{eq:normalisation}. Indeed, under certain assumptions (see~\ref{appendix:MCMC} for details), it is possible to define a stochastic process $\{R_0, R_1,\dots\}$, where $R_i$ is a discrete random variable describing the model's parameters $R$, that converges to the true posterior distribution $p(R|S,Y)$, by only knowing that $p(R|S,Y)\propto p(Y|S,R)p(R)$. MCMC provides different routines to simulate such a stochastic process and allows the collection of samples from the posterior without knowing explicitly its normalisation. In particular, in this work, we use the No-U-Turn Sampler (NUTS)~\cite{hoffman2014no}, which is a variant of the Hamiltonian Monte Carlo (HMC)~\cite{duane1987hybrid}.

One of the most important issues in MCMC is understanding whether the chain has converged or not to the posterior distribution~\cite{lynch2007introduction}. There are a few methods to evaluate convergence, but none of them is perfect, since without knowing the target distribution in advance it's impossible to guarantee that convergence has been reached. Even assuming that convergence is reached, it could take a lot of iterations depending on a number of factors, among which are the dimension of the parameters' space and the complexity of the target distribution. 
A large number of parameters negatively affects the use of MCMC also from a mere memory perspective, since to infer properties of the target distribution it is necessary to collect a relatively large number of samples.
Given the high dimensionality of the reservoir space, a straightforward application of MCMC is difficult.

We explored two possible solutions to overcome this limitation. 
The first one is to apply principal component analysis (PCA) to reduce the dimension of the reservoir state space before running MCMC. 
Besides reducing the computational burden in training the readout, this procedure also offers a regularisation that can improve the overall performance in the downstream task~\cite{lokse2017training}.

The second solution is to implement Stochastic Search Variable Selection (SSVS)~\cite{george1997approaches}, which was introduced to speed up convergence, following the intuition that more promising parts of the parameter space should be more likely and so MCMC would converge there faster. In a Bayesian regression problem, we consider a likelihood distribution of the form $p(Y|S,\beta,\sigma)=\mathcal{N}(S^T\beta,\sigma^2)$,\footnote{In this case we consider a linear readout $g(s;\beta)=s^T\beta$.} with $Y$ real-valued, and $S$ and $\beta$ $\mathbb{R}^p$-valued random variables, and we are interested in the posterior distribution of $\sigma$ and of the regression parameters $\beta$ that best fit the inputs $S$ to $Y$. 
The idea of SSVS is that some of the $\beta_i$'s might be negligible.
For this reason, the prior over all parameters is $p(\beta,\sigma)=p(\beta|\sigma)p(\sigma)$, where we chose $p(\sigma)$ to be a uniform distribution and $p(\beta|\sigma)$ is a \emph{shrinking} distribution~\cite{van2021bayesian}, inducing sparsity over $\beta$. 
As the shrinking distribution we chose the \emph{horseshoe} distribution~\cite{carvalho2009handling}, which introduces a local shrinkage parameter $\lambda_i$ and a global one $\tau$, distributed like follows
\begin{align}
    \beta_i &\sim \mathcal{N}(0, \lambda_i^2\tau^2) \\
    \lambda_i, \tau &\sim \mathcal{C}^+(0,1),
\end{align}
where $\mathcal{C}^+$ is a half-Cauchy distribution.

\subsection{Dropout}
\label{sec:dropout}

The dropout technique~\cite{srivastava2014dropout} can be used to easily morph the readout $g$, which in our case is an MLP, into a Bayesian model.
Consider an MLP model with just one hidden layer and without the bias term, for simplicity.
Let us denote $Q$ the dimension of the input, $K$ the number of hidden units and $D$ the output dimension. Dropout is applied by sampling two vectors $b_1$ and $b_2$ of dimensions $Q$ and $K$ from a Bernoulli distribution of parameter $p$ and by changing the action of the MLP as follows:
\begin{equation}
    \label{eq:dropout}
    y=\sigma(((s\circ b_1) W_1 \circ b_2)W_2),
\end{equation}
where $\circ$ stands for the Hadamard product and $\sigma$ is the activation function.

Dropout is commonly used during training as a regularisation technique~\cite{srivastava2014dropout}, but, when used during inference, it makes the MLP in~\eqref{eq:dropout} a fully Bayesian model~\cite{gal2015dropout} that can be applied iteratively to build an ensemble of values $\{y_i\}$, allowing to estimate uncertainties without the need to solve~\eqref{eq:normalisation}.

\subsection{Quantile Regression}
\label{sec:qr}

Another approach, other than using BNNs, is to change the training of $g$ so that for each reservoir state $s_t$ it outputs both the forecast $y_t$ and a measure $q$ of its uncertainty. In Quantile Regression (QR)~\cite{koenker1978regression} such $q$ can be one or more quantiles of choice. 
Given a probability distribution $P(Y|S)$, the quantile of level $\tau$ is that value $q_\tau(Y|S)$ such that $\prob(Y\leq q_\tau)=\tau$.
By simply redefining the loss function of a regression task, QR infers directly the quantile we are interested in.

In a standard regression problem, we are able to find the sample mean $\mathbf{E}[Y|S]=g(S;R)$ by minimising the sum of squared residuals, so we solve
\begin{equation}
    \argmin_R \sum_{t=1}^T\,\left(y_t-g(s_t;R)\right)^2.
\end{equation}
In QR the optimisation problem is modified as follows
\begin{equation}
    \argmin_{R(\tau)}\sum_{t=1}^T \,\rho_\tau\!\left(y_t-g(s_t;R(\tau))\right),
\end{equation}
where $\rho_\tau$ is called check loss (or pinball loss) and is defined as
\begin{equation}
    \rho_\tau(r)=\tau \max(r,0)+(1-\tau)\max(-r,0),
\end{equation}
and the regression parameters $R(\tau)$ will depend on the chosen quantile.

By performing QR for multiple values of $\tau$, we are able to infer a more detailed view of the distribution $P(Y|S)$, than just its mean value.

\subsection{Calibration}
\label{sec:calibration}

In all the above scenarios the model does not yield a simple real value, but it has a more structured output containing information about uncertainty. For this reason, we can't simply evaluate its performance by some accuracy measure alone. Indeed, the mean value or the median of the output can accurately reproduce $y_t$ and yet the output distribution can still be far off from the values spanned in the dataset.

\begin{figure}
     \centering
     \begin{subfigure}[b]{0.4\textwidth}
         \centering
         \includegraphics[width=\textwidth]{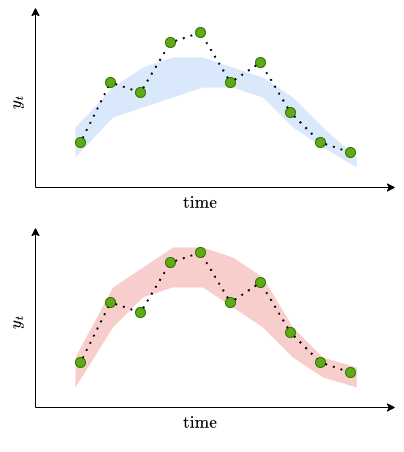}
         \caption{In the upper plot, the 90\% confidence interval fails to contain 90\% of the data points. Below, in red, is the corresponding calibrated interval.}
         \label{fig:traj-calib}
     \end{subfigure}
     \hfill
     \begin{subfigure}[b]{0.49\textwidth}
         \centering
         \includegraphics[width=\textwidth]{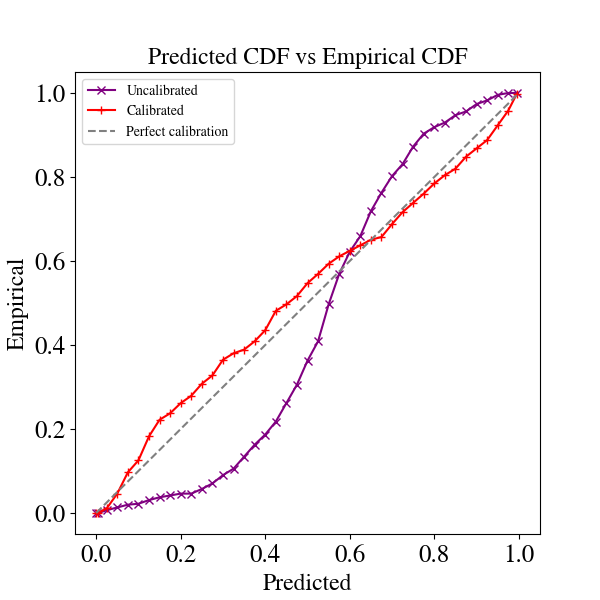}
         \caption{Example of the relationship between predicted $\tau$ and empirical CDF $\Tilde{\tau}$ before calibration (in purple), after calibration (in red) and the ideal case (dashed grey line).}
         \label{fig:calib-dataset}
     \end{subfigure}
     \caption{}
     \label{fig:calibration}
\end{figure}

The idea behind calibration is that if the output says that $I\subset\mathbb{R}$ is a $90\%$ confidence interval for $\tilde{y}_t$, then $y_t$ should fall in $I$ roughly $90\%$ of the time~\cite{pmlr-v80-kuleshov18a, gneiting2014probabilistic}. See Figure~\ref{fig:traj-calib} for an example.
Ideally, we need to check that the condition above holds for every possible confidence interval. 
In practice, we pick a set of $K$ quantile levels $\{\tau_i\}_{i=1}^K$, that span the interval $[0,1]$ and check the calibration on each of them. 
To do so we first need to compute for each time step the quantiles $q_{\tau,t}$, that is those values that satisfy
\begin{equation}
    \label{eq:quantiles}
    \tau=\int_{-\infty}^{q_{\tau,t}}\,p(y|s_t)\,dy=F_t(q_{\tau,t}) \qquad \implies \qquad q_{\tau,t}=F_t^{-1}(\tau),
\end{equation}
where $F_t$ is the cumulative density distribution (CDF). Notice that quantiles need to be computed from the sampled values of $\tilde{y}_t$ for VI, MCMC and dropout, while they are provided directly as output for QR.

Ideally, for each $\tau_i$, $y_t$ should be smaller than $q_{\tau_i,t}$ approximately with a frequency of $\tau_i$, that is
\begin{equation} \label{eq:calibration}
    \tilde{\tau}_i:=\frac{1}{T}\sum_{t=1}^T\,|\{y_t\,:\,y_t\leq q_{\tau_i,t}\}| \simeq \tau_i \qquad \text{for }i=1,\dots,K,
\end{equation}
where $|\cdot|$ represents the cardinality of a set. The left-hand side of~\eqref{eq:calibration}, that we called $\tilde{\tau}$, is the empirical cumulative density distribution, while on the right-hand side we have the theoretical one.

We can quantify how uncalibrated our model is by computing the calibration error
\begin{equation} \label{eq:cal-error}
    \text{cal}=\sum_{i=1}^K\,w_i(\tau_i-\tilde{\tau}_i)^2,
\end{equation}
where $w_i$ are weights that can be used to give less importance to some quantiles; in our case, we used $w_i=1$.

In order to calibrate a model, we use the method described in~\cite{pmlr-v80-kuleshov18a}: with a separate portion of the dataset, we create a recalibration dataset $\{(\tau_i,\tilde{\tau}_i)\}_{i=1}^K$ and we use it to train a calibration model $\mu:[0,1]\to [0,1]$, which is then used to correct the uncalibrated quantiles. Consider for example Figure~\ref{fig:calib-dataset}: the horizontal axis represents the predicted CDF $\tau$, while the vertical axis is the empirical CDF $\tilde{\tau}$. The purple points are the couples $\{(\tau_i,\tilde{\tau}_i)\}_{i=1}^K$ obtained from the test set before calibration, which mostly lay far from the dashed line where $\tau=\tilde{\tau}$. After applying the calibration model $\mu$, trained on the calibration set so that $\mu(\tau)$ approximates $\tilde{\tau}$, the results are much more calibrated, as shown in red in Figure~\ref{fig:calib-dataset}.

\section{Experimental framework}
\label{sec:experiments}

Our experiments aim at comparing in a time series forecasting task the performance of a probabilistic readout implemented with different uncertainty quantification techniques. The methods we employed are VI, MCMC (with and without the use of PCA to reduce the dimensionality of the reservoir state), SSVS, dropout and quantile regression. Our software implementation\footnote{Code available at \url{https://github.com/MicheleUIT/Probabilistic-load-forecasting-with-Reservoir-Computing}} makes use of PyTorch and the probabilistic programming library Pyro~\cite{bingham2018pyro}. The different methods are compared according to common criteria quantifying accuracy in the prediction, quality of the uncertainty and computational complexity.
For each method, we also discuss the simplicity of usage in terms of expertise required by the practitioner.

All experiments were run on NVIDIA RTX A6000 graphics cards.

\subsection{Datasets}
To carry out the experiments, two datasets of univariate time series have been used.

The first time series represents the electric load measured by Azienda Comunale Energia e Ambiente (ACEA), the company managing the electricity distribution in Rome. Data were collected every 10 minutes for 3 years of activity, from 2009 to 2011 \cite{bianchi2015short} with a precision up to the second decimal digit. From the original dataset, we took a point every 6, so to have an hourly resolution, and removed 11 weeks (being careful not to disrupt the pattern of the time series) due to a strong anomaly that would affect disproportionately the statistics of the training set compared to the calibration and test sets. The final dataset consists of 21048 time steps.

The second dataset (Spain) represents the daily electric demand in the Spanish market between 2014 and 2018, with a precision up to the fourth decimal digit.\footnote{Source: \url{https://www.esios.ree.es/en}} The dataset has a length of 1825 time steps.

In both cases, the training dataset was formed by the first 70\% of the time series, while the test and validation datasets both take half of the remaining 30\%.
We normalized the datasets by subtracting the mean and by dividing them by the standard deviation of the training set.

When working on time series forecasting, a crucial element to take into consideration is the presence of seasonalities. As expected, both ACEA and Spain datasets have a strong weekly seasonality, i.e.~$s=168$ for ACEA and $s=7$ for Spain. While seasonalities with longer lags are of course present, they are much less prominent.

The dataset will be of the form $\{(x_t,y_t)\}_{t=1}^T$, with $y_t=x_{t+h}$, where $h$ is the forecast horizon. The input $x_t$ is first processed by the reservoir to produce the states $s_t$, so the readout $g$ is trained with a dataset $\{(s_t,y_t)\}_{t=1}^T$. The choice of the forecast horizon $h$ is constrained by the particular application at hand, but it is also limited by the removed seasonality $s$: $h$ must be smaller or equal to $s$, or, if multiple seasonalities are removed, it must be smaller than the smallest seasonality. Indeed, after removing the seasonality, each data point in the dataset will become 
\begin{equation}
    \begin{aligned}
        \tilde{x}_t&=x_t-x_{t-s} \\
        \tilde{y}_t&=\tilde{x}_{t+h}=x_{t+h}-x_{t-s+h}.
    \end{aligned}
\end{equation}
To reconstruct the output is necessary to reintroduce the seasonality by undoing the differentiation.
In particular, the reconstructed value is $x_{t+h}=\tilde{x}_{t+h}+x_{t-s+h}$. If $h>s$ we can see that $t-s+h>t$, so $x_{t-s+h}$ is part of the forecast. The uncertainty of the forecaster will affect $x_{t+h}$ twice, from both terms.
This is an artefact stemming from the seasonal differencing procedure that inflates the uncertainty of the predictions.
If, instead, $h\leq s$, then $x_{t-s+h}$ is part of the input, so its value is known with certainty and it will not contribute to the uncertainty of the output. For this reason, we chose $h=24$ for ACEA and $h=1$ for Spain, i.e.~one day in both cases.

\subsection{Metrics}
To perform our analyses, we need metrics that
\begin{itemize}
    \item measure the capability of each method to produce accurate point forecasts and
    \item ascertain the quality of the produced uncertainty.
\end{itemize}
To perform a comparison, we necessitate these metrics to be applicable to each method.

As a measure of forecasting performance, we use the mean squared error (MSE) between the true value $y_t$ and the median of the forecast $g(s_t;R)$. Evaluating the uncertainty would not be very meaningful if the underlying forecast is not accurate, so the MSE is the main metric considered for hyperparameter selection. Referring to the median, instead of the average, allows the computation of the MSE homogeneously for all methods (in particular, with QR is not possible to compute the average), so it can be used to compare them.

Another important gauge of performance is provided by the training time used by the different methods. It must be noted that MCMC strictly speaking does not undergo a training process. The closest notion would be the time of convergence of the chains, but assessing when those converge is a difficult problem and an open research question~\cite{cowles1996markov, roy2020convergence}. For MCMC then we measure the total time elapsed while running the single chain.

To analyse the quality of the uncertainty estimate, we use four different metrics. The first one is the calibration error (cal), see equation~\eqref{eq:cal-error}, which quantifies how much the empirical cumulative distribution veers from the theoretical one (ideally, it would be zero). Then, we consider the width and the coverage of the 95\% confidence interval, which measure, respectively, how wide that confidence interval is and if it actually covers 95\% of the points. Namely, a smaller width reflects a less uncertain forecast, while the coverage should be close to $0.95$. We notice that the coverage is redundant since the information is already contained in the calibration error on each quantile. Nevertheless, we report also the coverage on the 95\% confidence interval as it is a popular performance measure that is adopted in many applications. The last metric is the mean continuous rank probability score (mCRPS), which measures the average distance between the cumulative distribution and the Heaviside function $\theta$,
\begin{align}
    &\text{CRPS}(F,x)=\int\,(F(y)-\theta(y-x))^2\,dy \label{eq:crps} \\
    &\text{mCRPS}(F)=\mathbb{E}_x[\text{CRPS}(F,x)].
\end{align}
The cumulative distribution $F$ in~\eqref{eq:crps} is known exactly when working with VI or MCMC, while is only known discretely with QR, through the predicted quantiles. In order to homogeneously compute the CRPS for all methods then, we approximate $F$ with a step function defined using quantiles, which can be easily computed in all cases. The integral can then be approximated with a sum. Referring to~\eqref{eq:quantiles}, we have
\begin{equation}
    F(y)=\tau_i \qquad \text{if } y\in (q_{\tau_i},q_{\tau_{i+1}}), \qquad \text{for $i=1,\dots,K$}.
\end{equation}
CRPS can be interpreted as a combined measure of both the accuracy (MSE) and calibration (cal).

\section{Results}
\label{sec:results}

\begin{table}
    \centering
    \small
    \[
    \begin{array}{l|l}
        \toprule
        \text{Hyperparameters} & \text{Searched values} \\
        \midrule
        \text{\# layers} & 1, 2, 3 \\
        \text{units per layer} & 8, 16, 32, 128, 256, 512 \\
        \text{activation} & \tanh,~\text{ReLU} \\
        \text{prior} & \mathcal{N}(0,1),~\mathcal{N}(0,10),~\text{Unif}(0,1),~\text{Unif}(0,10),~\text{horseshoe} \\
        \text{learning rate} & [0.0001, 0.1] \\
        \text{$p$ (dropout)} & [0.1, 0.9] \\
        \bottomrule
    \end{array}
    \]
    \caption{Schematic view of the possible values and intervals searched for each hyperparameter. The number of layers, units per layer and activation functions refer to the readout's configuration, so they are common to all methods. On the other hand, the prior refers only to VI, MCMC and SSVS; the dropout probability to the dropout method; the learning rate to VI, dropout and QR.} 
    \label{tab:hyper-param}
\end{table}

\begin{table}
\small
\centering
    \begin{tabular}{l|cccccc}
        \toprule
        \multirow{2}{*}{Parameters} & \multicolumn{6}{c}{Methods} \\
        \cmidrule{2-7}
         & VI & Dropout & MCMC & MCMC (PCA) & SSVS & QR \\
        \midrule
        \# layers & 1 & 2 & 1 & \makecell{\textcolor{blue}{3} \\\textcolor{OrangeRed}{2}} & 1 & 3 \\
        \cmidrule{1-7}
        units per layer & 512 & 512, 256 & 512 & \makecell{\textcolor{blue}{32, 16, 8} \\\textcolor{OrangeRed}{32, 16}} & 512 & 512, 256, 128 \\
        \cmidrule{1-7}
        activation & ReLU & ReLU & $\tanh$ & $\tanh$ & \makecell{\textcolor{blue}{$\tanh$} \\\textcolor{OrangeRed}{ReLU}} & \makecell{\textcolor{blue}{ReLU} \\\textcolor{OrangeRed}{$\tanh$}}  \\
        \cmidrule{1-7}
        prior $p(R)$ & \makecell{\textcolor{blue}{$\mathcal{N}(0,1)$} \\\textcolor{OrangeRed}{$\text{Unif}(0,1)$}} & N/A & \makecell{\textcolor{blue}{$\mathcal{N}(0,1)$} \\\textcolor{OrangeRed}{$\text{Unif}(0,1)$}} & \makecell{\textcolor{blue}{$\mathcal{N}(0,1)$} \\\textcolor{OrangeRed}{$\text{Unif}(0,1)$}} & horseshoe & N/A \\
        \cmidrule{1-7}
        prior $p(\Sigma)$ & $\text{Unif}(0,1)$ & N/A & $\text{Unif}(0,10)$ & $\text{Unif}(0,10)$ & $\text{Unif}(0,10)$ & N/A \\
        \cmidrule{1-7}
        learning rate & \makecell{\textcolor{blue}{$\sim 0.001$} \\\textcolor{OrangeRed}{$\sim 0.01$}} & \makecell{\textcolor{blue}{$\sim 0.007$} \\\textcolor{OrangeRed}{$\sim 0.01$}} & N/A & N/A & N/A & \makecell{\textcolor{blue}{$\sim 0.001$} \\\textcolor{OrangeRed}{$\sim 0.0006$}} \\
        \cmidrule{1-7}
        $p$ (dropout) & N/A & \makecell{\textcolor{blue}{$\sim 0.12$} \\\textcolor{OrangeRed}{$\sim 0.57$}} & N/A & N/A & N/A & N/A  \\
        \bottomrule
    \end{tabular}
    \caption{This table reports the optimal hyperparameters we used, for each method, to produce the results in Table~\ref{tab:results}. The optimal values are found by performing a grid search among all the possible combinations reported in Table~\ref{tab:hyper-param} and by selecting the best-performing one in terms of the accuracy of the forecast (MSE) obtained on the validation set. The colour blue refers to ACEA dataset, while red refers to Spain dataset. Those in black are common to both datasets. A hyperparameter is marked as not applicable (N/A) if it doesn't apply to that particular method.}
    \label{tab:chosen-param}
\end{table}

To ensure a fair comparison, the reservoir is generated once, configured with hyperparameters that satisfy the echo state property, and then used for all methods.
In addition, for each method, we optimised the hyperparameters using cross-validation based on grid search. The optimised hyperparameters involve both the structure of the MLP used for the readout $g$ and some method-specific features too. Table~\ref{tab:hyper-param} contains a list of all the possible values for each hyperparameter, while Table~\ref{tab:chosen-param} reports the optimal hyperparameter configuration for each method on the two analysed datasets.

\begin{figure}
     \centering
     \begin{subfigure}[b]{0.3\textwidth}
         \centering
         \includegraphics[width=\textwidth]{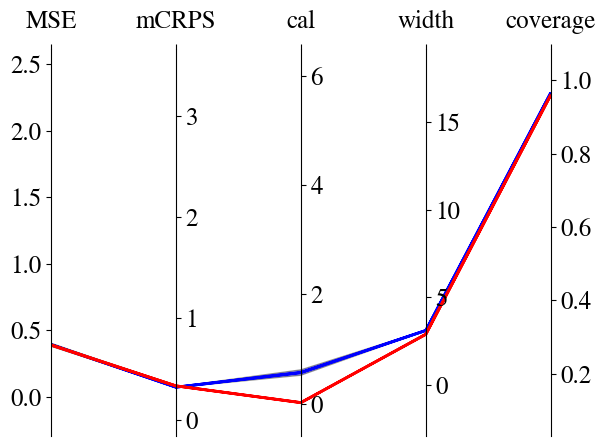}
         \caption{VI}
         \label{fig:calib-svi-acea}
     \end{subfigure}
     \hfill
     \begin{subfigure}[b]{0.3\textwidth}
         \centering
         \includegraphics[width=\textwidth]{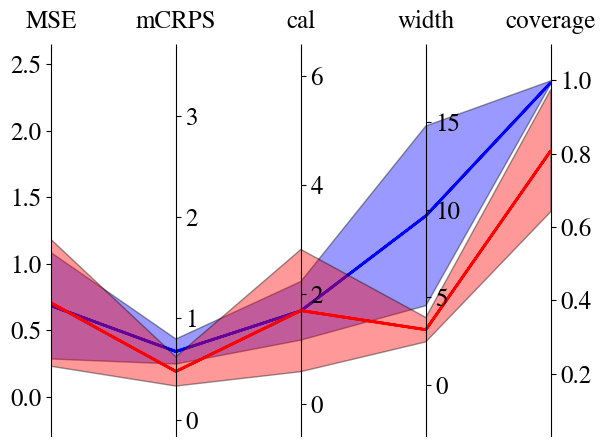}
         \caption{MCMC without PCA}
     \end{subfigure}
     \hfill
     \begin{subfigure}[b]{0.3\textwidth}
         \centering
         \includegraphics[width=\textwidth]{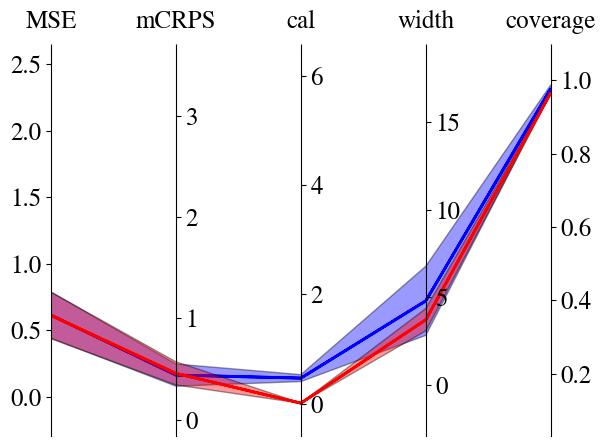}
         \caption{MCMC with PCA}
     \end{subfigure}
     
     \begin{subfigure}[b]{0.3\textwidth}
         \centering
         \includegraphics[width=\textwidth]{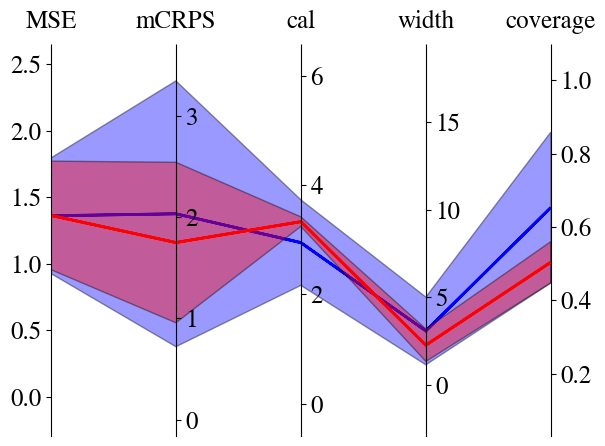}
         \caption{SSVS}
     \end{subfigure}
     \hfill
     \begin{subfigure}[b]{0.3\textwidth}
         \centering
         \includegraphics[width=\textwidth]{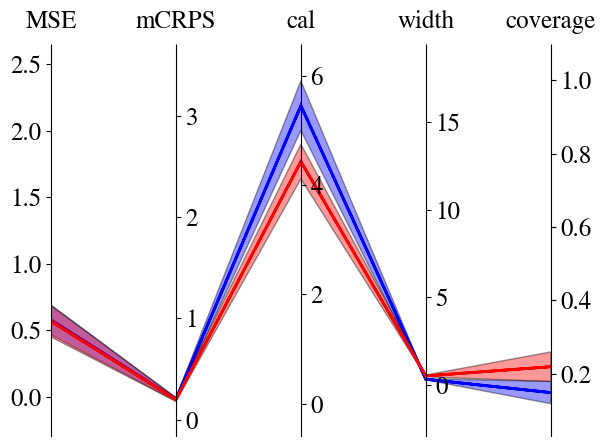}
         \caption{Dropout}
         \label{fig:dropout-acea}
     \end{subfigure}
     \hfill
     \begin{subfigure}[b]{0.3\textwidth}
         \centering
         \includegraphics[width=\textwidth]{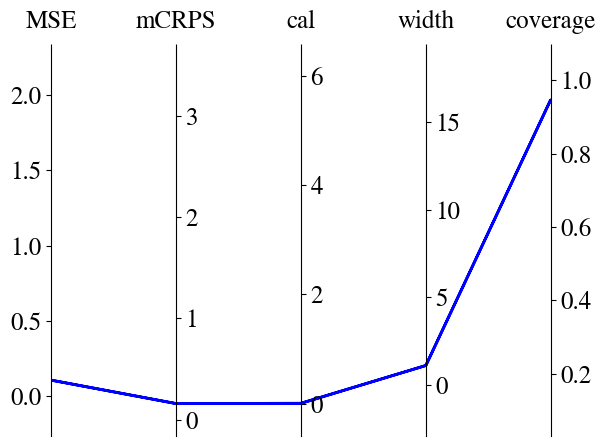}
         \caption{QR}
         \label{fig:qr-acea}
     \end{subfigure}
        \caption{Visual representation of the results in Table~\ref{tab:results} for ACEA dataset, before (in blue) and after calibration (in red). Notice that calibration is not performed on QR because the predicted CDF is discontinuous.}
        \label{fig:calib-acea}
\end{figure}

\begin{figure}
     \centering
     \begin{subfigure}[b]{0.3\textwidth}
         \centering
         \includegraphics[width=\textwidth]{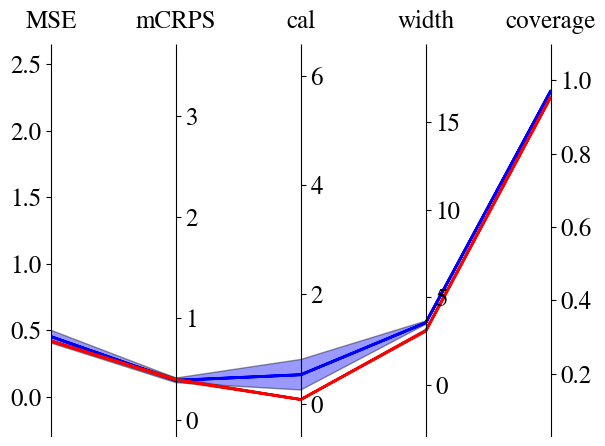}
         \caption{VI}
     \end{subfigure}
     \hfill
     \begin{subfigure}[b]{0.3\textwidth}
         \centering
         \includegraphics[width=\textwidth]{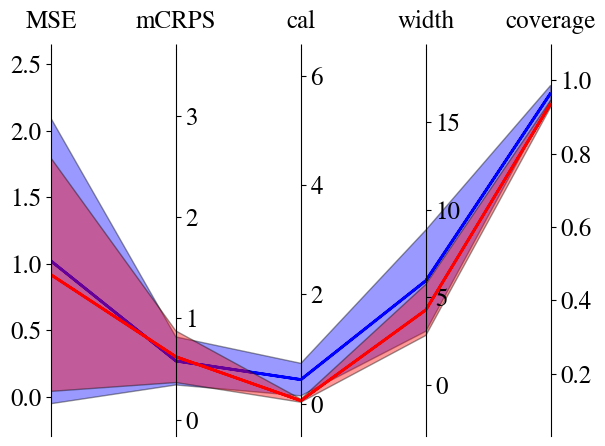}
         \caption{MCMC without PCA}
     \end{subfigure}
     \hfill
     \begin{subfigure}[b]{0.3\textwidth}
         \centering
         \includegraphics[width=\textwidth]{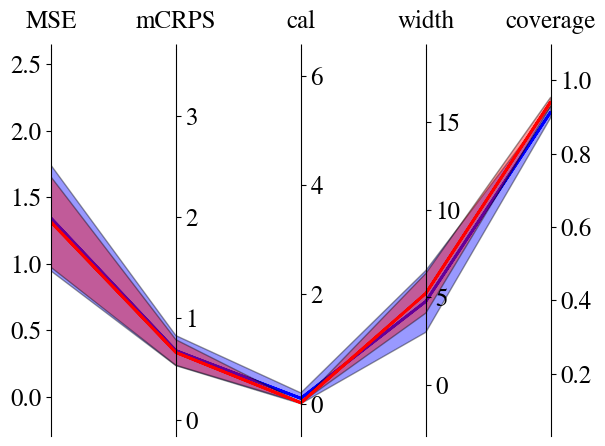}
         \caption{MCMC with PCA}
     \end{subfigure}
     
     \begin{subfigure}[b]{0.3\textwidth}
         \centering
         \includegraphics[width=\textwidth]{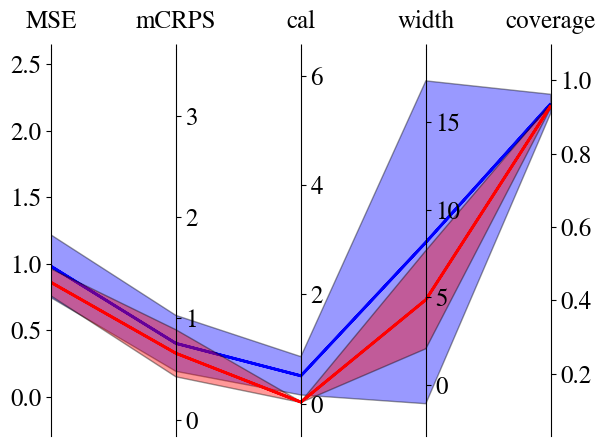}
         \caption{SSVS}
     \end{subfigure}
     \hfill
     \begin{subfigure}[b]{0.3\textwidth}
         \centering
         \includegraphics[width=\textwidth]{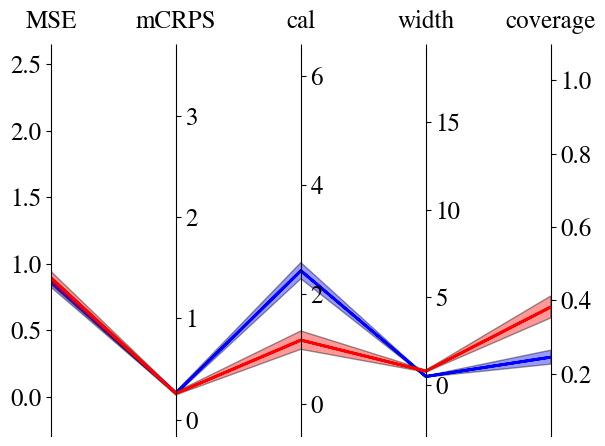}
         \caption{Dropout}
         \label{fig:dropout-spain}
     \end{subfigure}
     \hfill
     \begin{subfigure}[b]{0.3\textwidth}
         \centering
         \includegraphics[width=\textwidth]{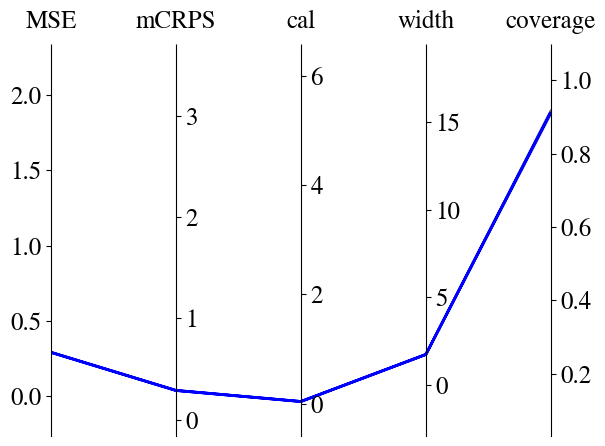}
         \caption{QR}
         \label{fig:qr-spain}
     \end{subfigure}
        \caption{Visual representation of the results in Table~\ref{tab:results} for Spain dataset, before (in blue) and after calibration (in red). Notice that calibration is not performed on QR because the predicted CDF is discontinuous.}
        \label{fig:calib-spain}
\end{figure}

Once the best-performing hyperparameters have been determined, we perform 10 runs for each experiment and collect the resulting means and standard deviations in Table~\ref{tab:results}: it shows the performance of each method in terms of the computational burden (training time), the accuracy of the forecast (MSE) and the robustness of the uncertainty (cal, width, coverage and mCRPS). Plots in Figures~\ref{fig:calib-acea} and~\ref{fig:calib-spain} present, tersely, the same results, with the additional effect of calibration on each metric.
Specifically, the course of a blue streak moves through the metrics before calibration, with a solid line representing the mean and the shaded area the standard deviation, while the red band depicts those same metrics after calibration occurred.

\begin{table}
    \centering
    \footnotesize
    \[
    \begin{array}{cc|ccccc|c}
        \toprule
        \text{Method} & \text{Dataset} & \text{width (95\%)} & \text{coverage (95\%)} & \text{cal} & \text{mCRPS} &\text{MSE} & \text{training time (\unit{\second})}  \\
        \midrule
        \multirow[c]{2}{*}{\text{VI}}
        &\text{\textcolor{blue}{ACEA}} & 3.11\pm0.01 & 0.97\pm0.00 & 0.57\pm0.05 & 0.32\pm0.00 &0.39\pm0.01 & 10.41\pm0.16  \\
        &\text{\textcolor{OrangeRed}{Spain}} & 3.55\pm0.09 & 0.97\pm0.00 & 0.53\pm0.28 & 0.39\pm0.02 &0.45\pm0.05 & 2.78\pm0.17  \\
        \cmidrule{1-8}
        \multirow[c]{2}{*}{\text{Dropout}}
        &\text{\textcolor{blue}{ACEA}} & 0.32\pm0.04 & 0.15\pm0.03 & 5.46\pm0.45 & 0.20\pm0.02 &0.58\pm0.11 & {\color{blue}\mathbf{0.10\pm0.14}}  \\
        &\text{\textcolor{OrangeRed}{Spain}} & 0.48\pm0.03 & 0.25\pm0.02 & 2.43\pm0.15 & {\color{OrangeRed}\mathbf{0.26\pm0.02}} &0.86\pm0.04 & {\color{OrangeRed}\mathbf{0.08\pm0.14}}  \\
        \cmidrule{1-8}
        \multirow[c]{2}{*}{\text{MCMC}}
        &\text{\textcolor{blue}{ACEA}} & 9.64\pm5.12 & 0.99\pm0.00 & 1.70\pm0.54 & 0.67\pm0.12 &0.69\pm0.40 & (3\pm4)\times 10^3  \\
        &\text{\textcolor{OrangeRed}{Spain}} & 5.95\pm2.89 & 0.97\pm0.02 & 0.44\pm0.30 & 0.58\pm0.24 &1.02\pm1.07 & (3\pm2)\times 10^3  \\
        \cmidrule{1-8}
        \multirow[c]{2}{*}{\text{\makecell{MCMC \\ (PCA)}}}
        &\text{\textcolor{blue}{ACEA}} & 4.80\pm1.97 & 0.98\pm0.01 & 0.47\pm0.06 & 0.44\pm0.11 &0.62\pm0.17 & (6\pm6)\times 10^3  \\
        &\text{\textcolor{OrangeRed}{Spain}} & 4.77\pm1.76 & 0.91\pm0.02 & 0.10\pm0.10 & 0.68\pm0.15 &1.34\pm0.40 & (8\pm5)\times 10^3  \\
        \cmidrule{1-8}
        \multirow[c]{2}{*}{\text{SSVS}}
        &\text{\textcolor{blue}{ACEA}} & 3.08\pm1.93 & 0.65\pm0.21 & 2.95\pm0.78 & 2.03\pm1.31 &1.36\pm0.43 & (17\pm5)\times 10^3  \\
        &\text{\textcolor{OrangeRed}{Spain}} & 8.13\pm9.20 & {\color{OrangeRed}\mathbf{0.94\pm0.03}} & 0.51\pm0.35 & 0.75\pm0.28 &0.98\pm0.24 & (13\pm7)\times 10^3  \\
        \cmidrule{1-8}
        \multirow[c]{2}{*}{\text{QR}}
        &\text{\textcolor{blue}{ACEA}} & 1.11\pm0.03 & {\color{blue}\mathbf{0.95\pm0.00}} & {\color{blue}\mathbf{0.00\pm0.00}} & {\color{blue}\mathbf{0.16\pm0.00}} & {\color{blue}\mathbf{0.11\pm0.00}} & 2.92\pm0.23  \\
        &\text{\textcolor{OrangeRed}{Spain}} & 1.74\pm0.05 & 0.91\pm0.01 & {\color{OrangeRed}\mathbf{0.04\pm0.01}} & 0.29\pm0.01 & {\color{OrangeRed}\mathbf{0.29\pm0.00}} & 2.69\pm0.14  \\
        \bottomrule
    \end{array}
    \]
    \caption{Results are obtained by averaging 10 runs after selecting the best-performing hyperparameters of each method. The best values are in bold, with the colours encoding the datasets.}
    \label{tab:results}
\end{table}

Figures~\ref{fig:calib-acea} and~\ref{fig:calib-spain} show that the effect of calibration is different in each method. For example, Figure~\ref{fig:calib-svi-acea} shows that, for the ACEA dataset, VI takes advantage of the calibration process: the MSE gets a bit smaller even though not in a statistically significant way (from $0.394\pm0.010$ to $0.391\pm0.009$), but the calibration error drops almost to zero (from $0.568\pm0.049$ to $0.017\pm0.004$), with better coverage of the 95\% confidence interval (from $0.967\pm0.001$ to $0.961\pm0.003$).

Post-calibration bands for QR are not reported in Figures~\ref{fig:calib-acea} and~\ref{fig:calib-spain}. 
Since QR encompasses a discrete number of quantiles ($K=42$) and the predicted CDF is discontinuous, the calibration model $\mu$ could potentially point to quantiles that were not computed.
Indeed, the 95\% CI is defined by $(q_{\tau'},q_{\tau''})$, where $\tau'=0.025$ and $\tau''=0.975$, but, after calibrating, $\mu(\tau')$ and $\mu(\tau'')$ will, in general, be different, so the corresponding CI will not refer to 95\% anymore. All the other methods provide a distribution as output, so it is possible to compute new quantiles $q_{\tilde\tau'}$ and $q_{\tilde\tau''}$ corresponding to the calibrated CI. QR instead outputs the quantiles directly, so if the calibrated quantiles were not computed beforehand, they can not be obtained afterwards (it would be necessary to re-train the model anew, but it will be uncalibrated again).

\begin{figure}
     \centering
     \begin{subfigure}[b]{0.48\textwidth}
         \centering
         \includegraphics[width=\textwidth]{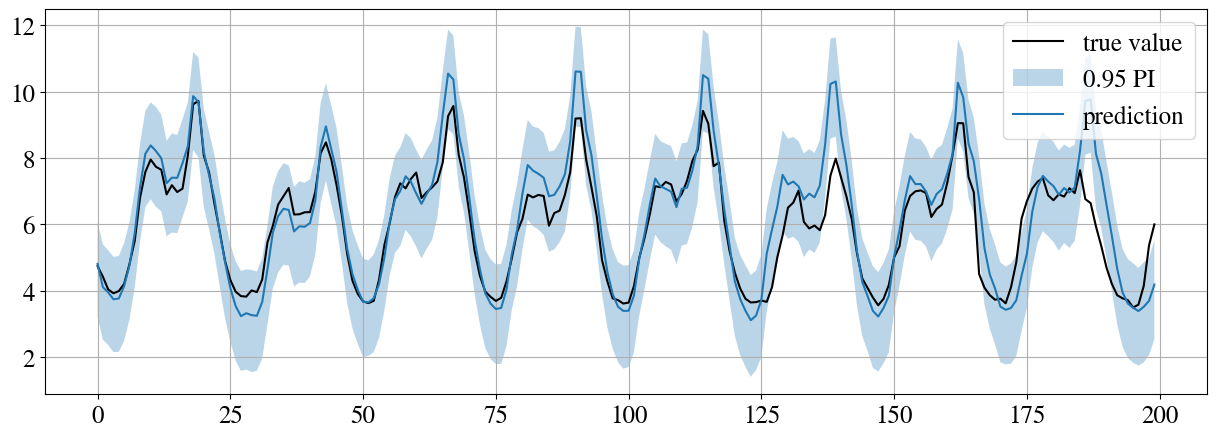}
         \caption{VI on ACEA.}
     \end{subfigure}
     \hfill
     \begin{subfigure}[b]{0.48\textwidth}
         \centering
         \includegraphics[width=\textwidth]{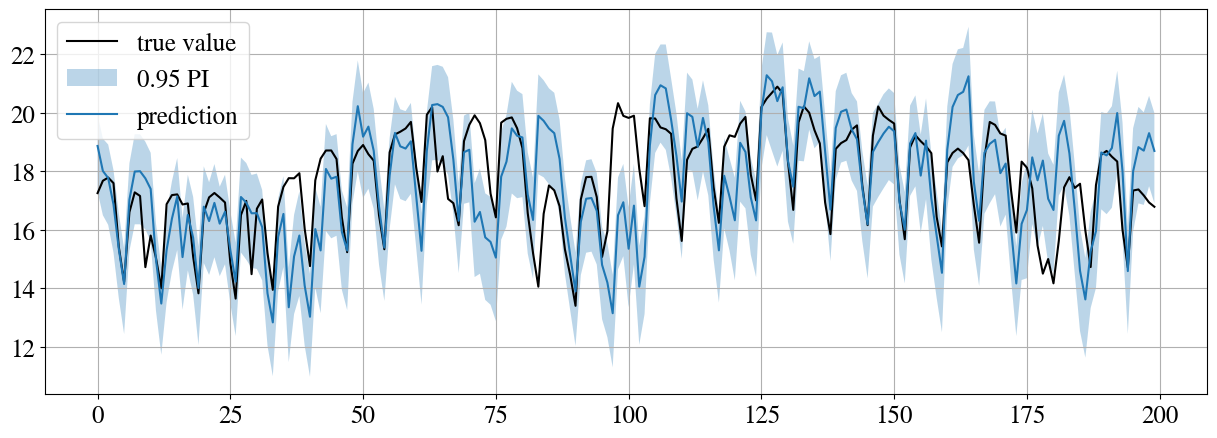}
         \caption{VI on Spain.}
     \end{subfigure}
     
     \centering
     \begin{subfigure}[b]{0.48\textwidth}
         \centering
         \includegraphics[width=\textwidth]{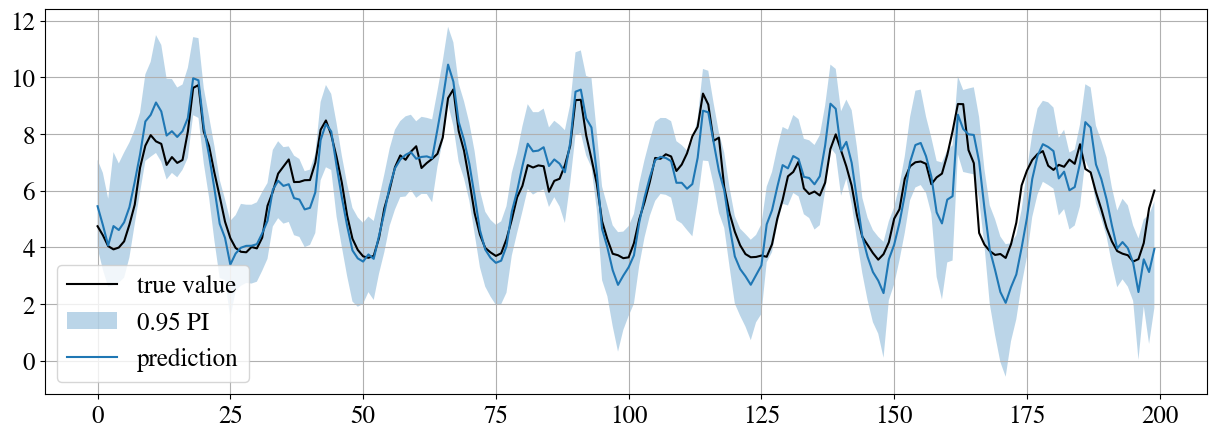}
         \caption{MCMC on ACEA.}
     \end{subfigure}
     \hfill
     \begin{subfigure}[b]{0.48\textwidth}
         \centering
         \includegraphics[width=\textwidth]{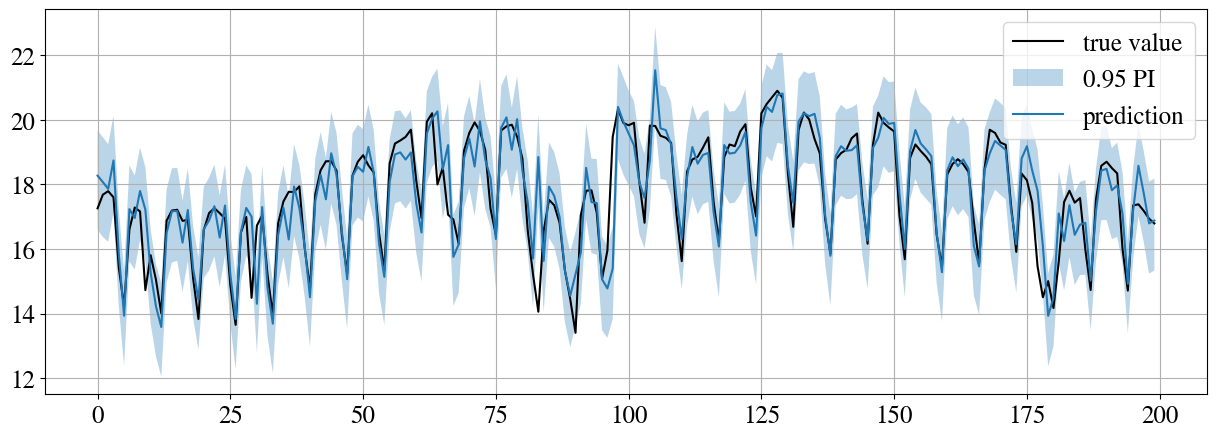}
         \caption{MCMC on Spain.}
     \end{subfigure}

    \centering
     \begin{subfigure}[b]{0.48\textwidth}
         \centering
         \includegraphics[width=\textwidth]{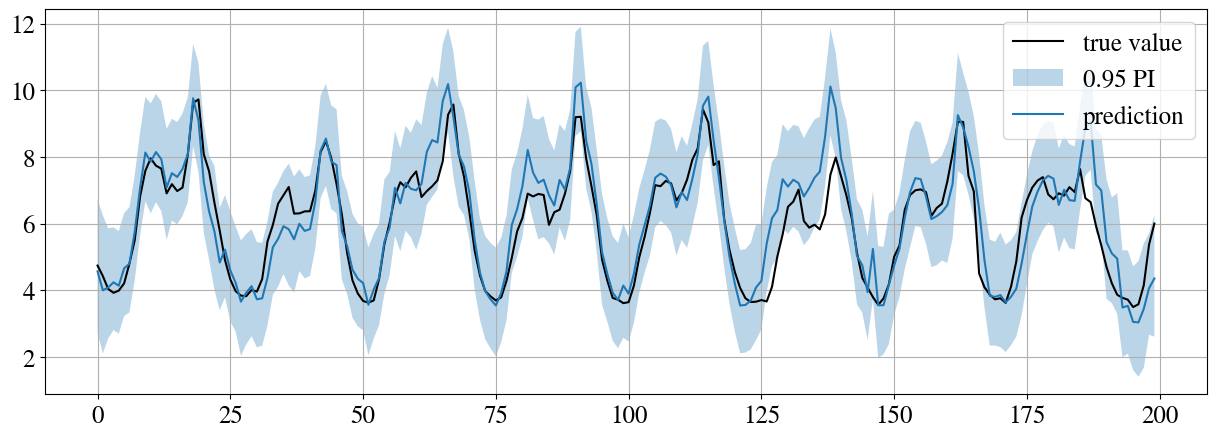}
         \caption{MCMC (PCA) on ACEA.}
     \end{subfigure}
     \hfill
     \begin{subfigure}[b]{0.48\textwidth}
         \centering
         \includegraphics[width=\textwidth]{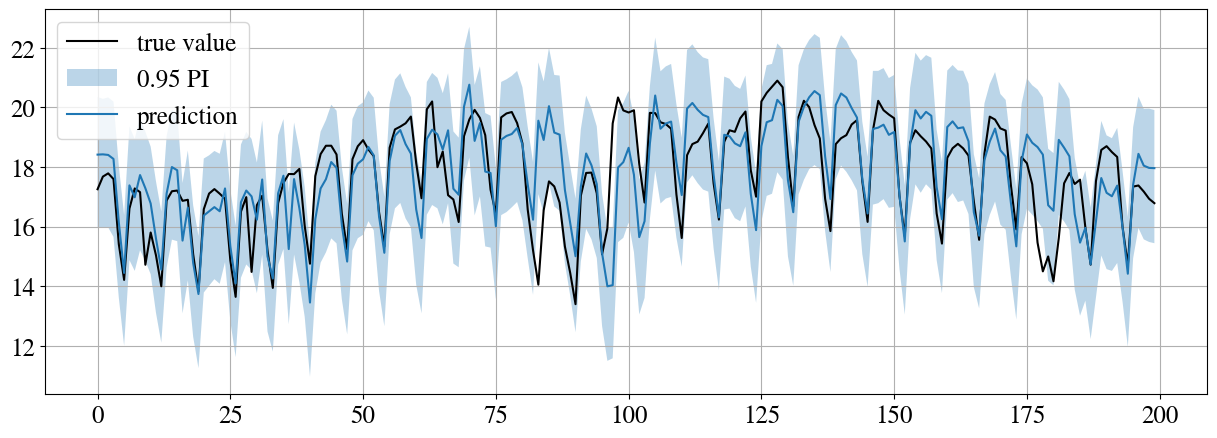}
         \caption{MCMC (PCA) on Spain.}
     \end{subfigure}

    \centering
     \begin{subfigure}[b]{0.48\textwidth}
         \centering
         \includegraphics[width=\textwidth]{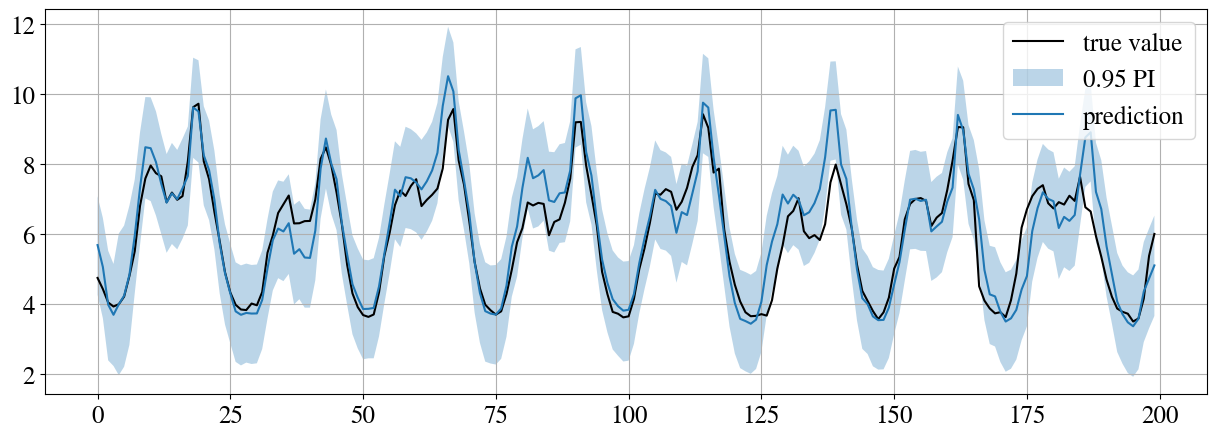}
         \caption{SSVS on ACEA.}
     \end{subfigure}
     \hfill
     \begin{subfigure}[b]{0.48\textwidth}
         \centering
         \includegraphics[width=\textwidth]{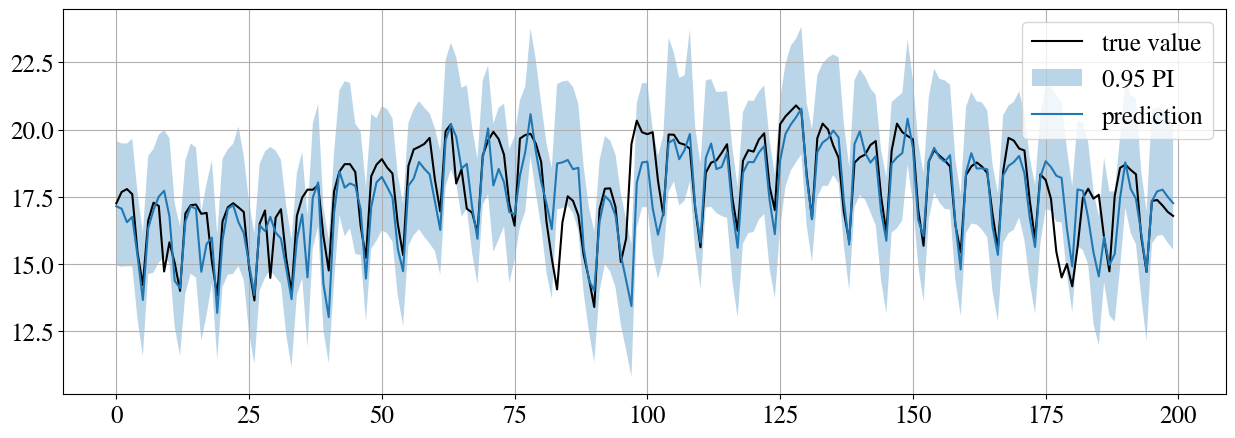}
         \caption{SSVS on Spain.}
     \end{subfigure}

    \centering
     \begin{subfigure}[b]{0.48\textwidth}
         \centering
         \includegraphics[width=\textwidth]{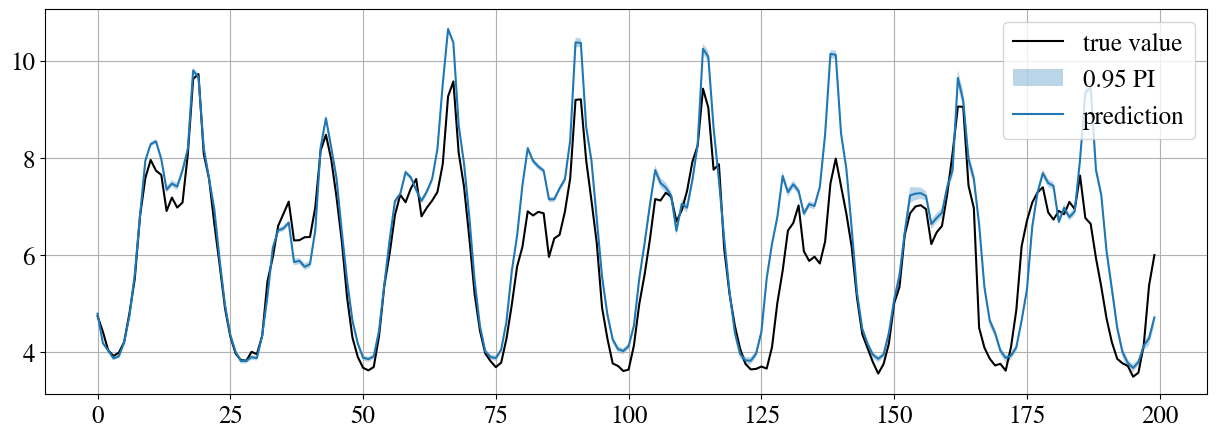}
         \caption{Dropout on ACEA.}
         \label{fig:forecast-dropout-acea}
     \end{subfigure}
     \hfill
     \begin{subfigure}[b]{0.48\textwidth}
         \centering
         \includegraphics[width=\textwidth]{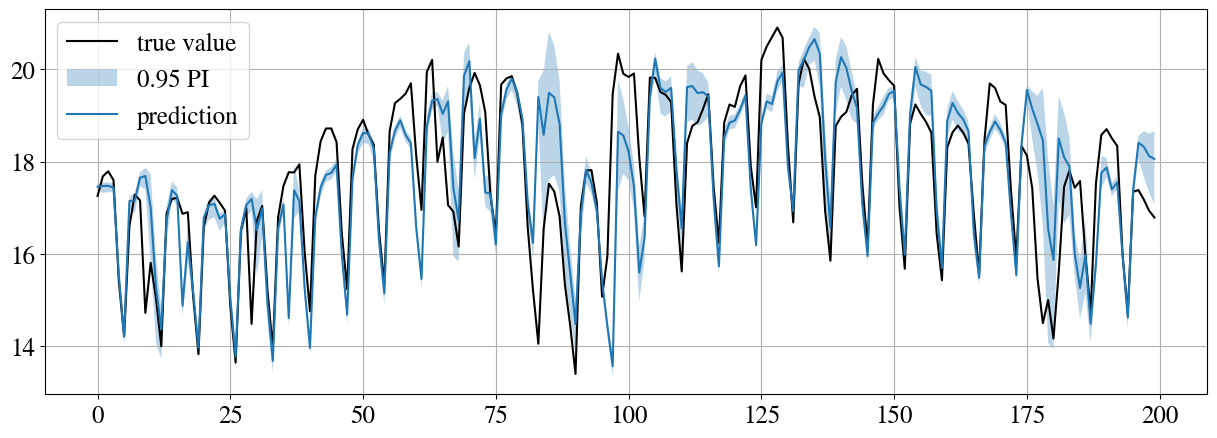}
         \caption{Dropout on Spain.}
         \label{fig:forecast-dropout-spain}
     \end{subfigure}

    \centering
     \begin{subfigure}[b]{0.48\textwidth}
         \centering
         \includegraphics[width=\textwidth]{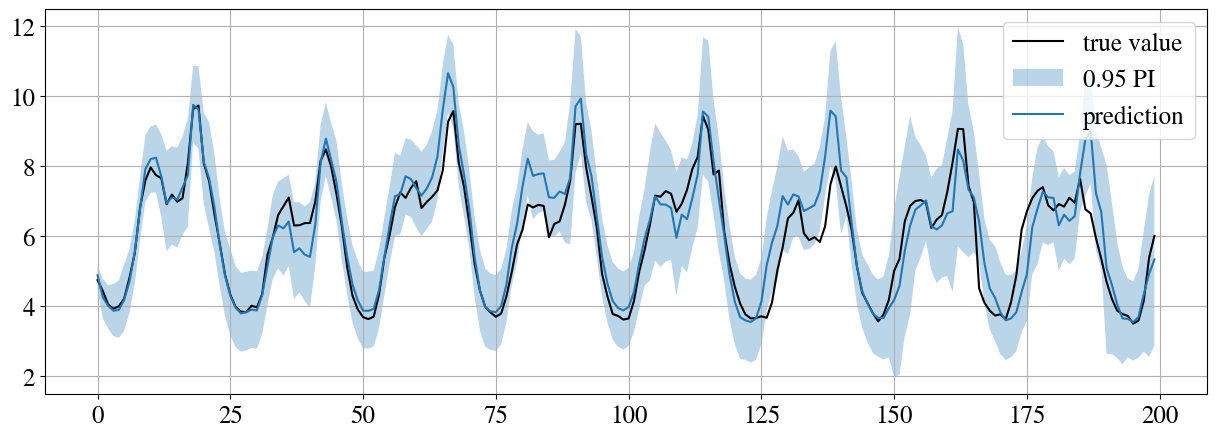}
         \caption{QR on ACEA.}
     \end{subfigure}
     \hfill
     \begin{subfigure}[b]{0.48\textwidth}
         \centering
         \includegraphics[width=\textwidth]{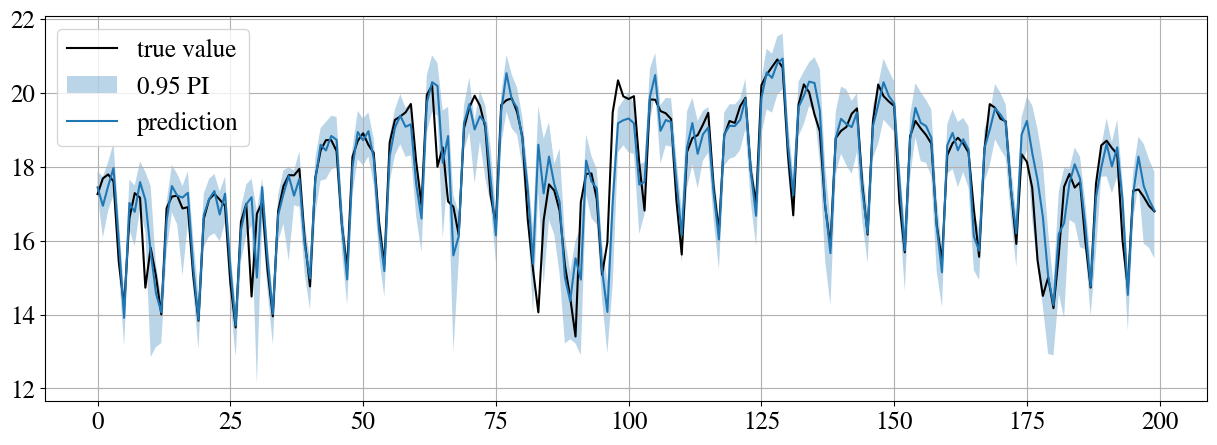}
         \caption{QR on Spain.}
     \end{subfigure}
     
    \caption{Examples of forecast trajectories for both datasets using all the discussed methods, after calibration. The black line represents the true value, the solid blue line is the median of the prediction and the shaded area is the 95\% confidence interval.}
    \label{fig:forecast}
\end{figure}

Figure~\ref{fig:forecast} shows some examples of forecasts produced by each method for both datasets, after calibration (except for QR).

\begin{figure}
     \centering
     \begin{subfigure}[b]{0.45\textwidth}
         \centering
         \includegraphics[width=\textwidth]{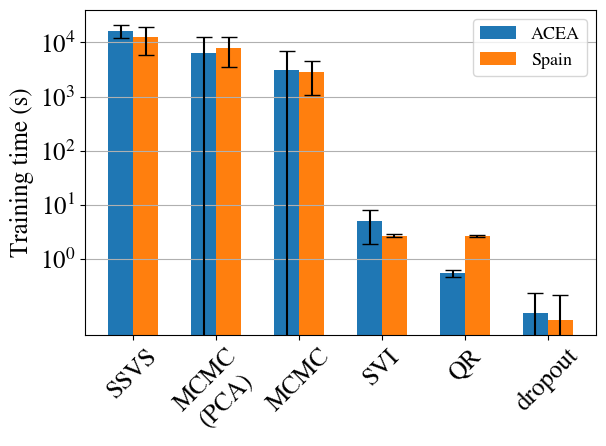}
         \caption{Comparison of the training times on the different datasets using a logarithmic scale.}
         \label{fig:train-times}
     \end{subfigure}
     \hfill
     \begin{subfigure}[b]{0.45\textwidth}
         \centering
         \includegraphics[width=\textwidth]{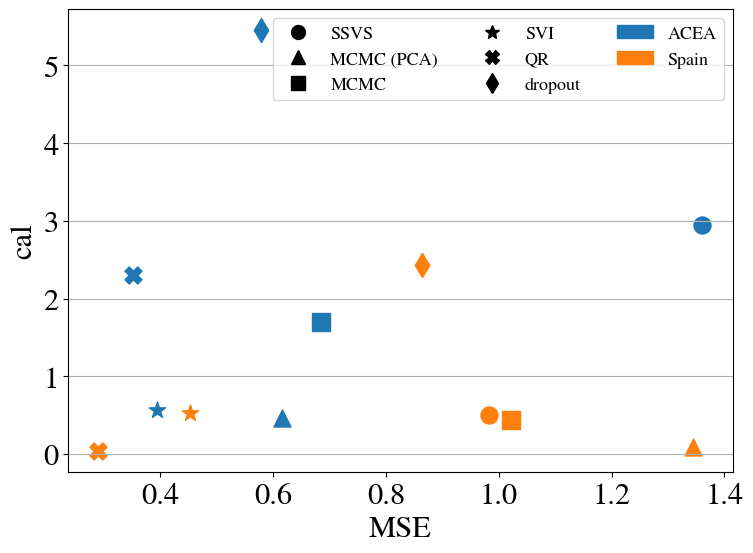}
         \caption{Calibration error against MSE. Bottom-left indicates the best performance, top-right the worst.}
        \label{fig:train-calib}
     \end{subfigure}
        \caption{}
\end{figure}

By looking at Table~\ref{tab:results} and the plots in Figures~\ref{fig:calib-acea},~\ref{fig:calib-spain} and~\ref{fig:forecast}, we notice that it is difficult, at first glance, to identify which is the best-performing method, since there are different methods that achieve the best performance, according to some metric, on each dataset. This underlines the complexity and possible ambiguities in evaluating the performance in probabilistic forecasting.
Nevertheless, there are important considerations that can be made, which are reported in the following.
\begin{itemize}
    \item The most pronounced difference regards training times, where methods based on MCMC show times from 2 to 4 orders of magnitude longer than other methods. The difference is further stressed in Figure~\ref{fig:train-times}. Although it is improper to talk about training time with MCMC, it does not change the fact that the MCMC method is extremely expensive and impractical to use in this scenario. Even by compressing the dimensionality of the reservoir states via PCA or by using SSVS, the computing times do not decrease.
    \item Overall, QR is the method that achieves better performance, especially on ACEA dataset. Even if post hoc calibration cannot be applied for the reasons discussed above, QR still achieves the lowest calibration error.
    \item All methods based on MCMC behave generally worse according to all metrics. In particular, the MSE in MCMC is higher than in other methods. The standard deviation of the results is also larger than other methods: this may be due to poor convergence of the chains.
    \item The fastest technique is dropout, which also achieves an MSE comparable with the other approaches. However, the calibration error, out of the box, is definitely higher due to the extremely narrow confidence intervals that yield poor coverage.
    Even after calibration, the uncertainty estimation does not improve by much as we can see from Figures~\ref{fig:dropout-acea} and~\ref{fig:dropout-spain}). 
    The poor performance in terms of uncertainty quantification is apparent by looking at Figures~\ref{fig:forecast-dropout-acea} and  \ref{fig:forecast-dropout-spain}: on both datasets, the uncertainty is generally very low, meaning that the randomness induced by dropout is not strong enough. With a higher value of $p$ the predicted uncertainty improves, but the MSE gets worse.
    \item Figure~\ref{fig:train-calib} depicts the relationship between uncertainty (cal) and forecast accuracy (MSE), with datasets shown with different colours. The better-performing methods are QR and VI:
    QR achieves the best performance on the Spain dataset, while VI is the best-performing method on ACEA and the second-best on Spain.
\end{itemize}

Overall, the approach that prevails when jointly considering the low training times, the MSE and the calibration error is QR. Its performance in terms of prediction accuracy and uncertainty estimation is comparable to those of VI, but its computational complexity is significantly lower.
Another important advantage of QR is the simplicity of implementing it: it is sufficient to replace the standard MSE loss used for training a regression model with the pinball loss, making QR a straightforward choice to be used in machine learning applications.
On the other hand, Bayesian approaches such as MCMC and VI rely on additional software libraries for probabilistic programming that adds an additional layer of complexity in terms of implementation.
Additionally, while Bayesian inference offers more flexibility thanks to the possibility of specifying the data distribution and introducing prior information in the model, it also poses additional challenges.
Indeed, the user is given the additional task of identifying the optimal distributions and the most meaningful priors, which is often not straightforward in many real-world applications, in addition to practical challenges such as checking the convergence of the chains in MCMC.

\begin{table}
    \centering
    \footnotesize
    \[
    \begin{array}{cc|ccccc|c}
        \toprule
        \text{Method} & \text{Dataset} & \text{width (95\%)} & \text{coverage (95\%)} & \text{cal} & \text{mCRPS} &\text{MSE} & \text{training time (\unit{\second})}  \\
        \midrule
        \multirow[c]{2}{*}{\text{ARIMA}}
        &\text{\textcolor{blue}{ACEA}} & 3.08 & 0.94 & 0.30 & 0.35 & 0.77 & 6.7\times 10^3  \\
        &\text{\textcolor{OrangeRed}{Spain}} & 2.35 & 0.84 & 0.05 & 0.39 & 0.95 & 30.44  \\
        \cmidrule{1-8}
        \multirow[c]{2}{*}{\text{DeepAR}}
        &\text{\textcolor{blue}{ACEA}} & 2.00\pm0.25 & 0.93\pm0.03 & 0.37\pm0.53 & 0.26\pm0.53 & 0.21\pm0.03 & (1.8\pm0.5)\!\times\! 10^3  \\
        &\text{\textcolor{OrangeRed}{Spain}} & 2.29\pm0.56 & 0.91\pm0.04 & 0.22\pm0.34 & 0.36\pm0.08 & 0.58\pm0.12 & 196.90\pm73.21  \\
        \bottomrule
    \end{array}
    \]
    \caption{Results for DeepAR are obtained by averaging 10 runs after selecting the best-performing hyperparameters for each dataset. On the other hand, since there is no stochasticity in its fitting procedure, ARIMA is executed only once and we do not report standard deviation in the results.}
    \label{tab:baseline}
\end{table}

\paragraph{Comparison with baselines}
To understand how the reservoir-based methods perform with respect to other baselines, we compare them against two popular methods for probabilistic time-series forecasting, ARIMA~\cite{box2015time} and DeepAR~\cite{salinas2020deepar}.
To determine the ARIMA model we relied on Auto-ARIMA from the pmdarima library\footnote{Source: \url{https://alkaline-ml.com/pmdarima/index.html}}, which identified $\text{ARIMA}(3,0,1)$ and $\text{ARIMA}(9,0,0)$ as the best-performing models for ACEA and Spain, respectively.
For DeepAR we followed the same hyperparameters' search procedure as for the reservoir-based models.
The best-performing DeepAR model has 2 RNN layers with a hidden size of 32 for ACEA and 8 RNN layers with a hidden size of 16 for Spain.
Table~\ref{tab:baseline} presents the same metrics obtained by ARIMA and DeepAR on the two datasets. 
As expected, the performance of ARIMA according to the different metrics is worse than most reservoir-based methods.
The accuracy of DeepAR is comparable with the other methods but with a much longer training time.
Finally, we notice how in both ARIMA and DeepAR the calibration error is aligned with those from the other methods, and that training times are much more sensitive than reservoir methods to the size of the dataset.

\section{Conclusion}
In the present paper, we delved into the task of time series forecasting in an energy analytics setting. As the core model, we deployed an Echo State Network, which, via its reservoir's dynamics, is able to unravel the possibly complex inner dependencies of the input time series and embed it in its state space. We argued that a point estimate, even if accurate, is often insufficient for policymakers in high-risk contexts like power market management, so it needs to be accompanied by a measure of how likely the forecast is. For this reason, we combined the Echo State Network with popular methods for uncertainty quantification. Specifically, we considered deterministic methods, like quantile regression, and Bayesian approaches, like dropout, variational inference and Markov chain Monte Carlo. 
The goal of our work was first to adapt them to handle the high-dimensional states of the reservoir and then compare them, by identifying their advantages and disadvantages.

Our experiments show that advanced methodologies like VI and MCMC, although they provide fine control over parameters' priors, do not substantiate a significant advantage in terms of forecasting accuracy or calibration, in fact proving to be demanding in terms of computational resources.
Quantile regression, on the other hand, does not seem to be affected by the high dimensionality of the reservoir states, emerging as fast to train, while achieving performances on par with the other methods on other indicators, without even using a post hoc calibration step.

\paragraph{Acknowledgments}
    The authors gratefully acknowledge NVIDIA Corporation for the donation of two RTX A6000 that were used in this project.

\appendix
\section{Appendix}

\subsection{Markov Chain Monte Carlo} \label{appendix:MCMC}

For simplicity, here we summarise MCMC theory for Markov chains with discrete state space, but it can be extended to continuous state space, which is what will is used in the present work.

A Markov Chain is a stochastic process $X=\{X_0, X_1,\dots\}$, where $X_i$ is a discrete random variable, that satisfies the Markov property
\begin{equation*}
	P(X_{n+1}=k|X_n=k_n,X_{n-1}=k_{n-1},\dots,X_1=k_1)=P(X_{n+1}=k|X_n=k_n),
\end{equation*}
i.e., the $(n+1)$-th state of the system depends only on the $n$-th one.
If we denote as $S(n)$ the distribution of states at time $n$, i.e., $S_i(n)=P(X_n=i)$, then we can write the distribution at time step $n+1$ as $S(n+1)=S(n)Q(n)$, where $Q_{ij}(n)=P(X_{n+1}=j|X_n=i)$ is the transition probability.
A Markov chain is homogeneous if the transition probability $Q_{ij}$ doesn't depend on time $n$, i.e.,
\begin{equation*}
	P(X_{n+1}=j|X_n=i)=P(X_{1}=j|X_0=i).
\end{equation*}
In this case, we can construct the whole trajectory of the chain states by iteratively applying $Q$ and it may saturate at a stationary distribution $S_i^*=\sum_j S_j^*Q_{ji}$, regardless of the initial state. A sufficient condition for a Markov chain with distribution $S$ to reach a stationary distribution is when $S$ satisfies detailed balance
\begin{equation*}
    S_iQ_{ij}=S_jQ_{ji} \quad \forall i,j.
\end{equation*}

The idea behind MCMC is to define a Markov chain that satisfies detailed balance such that it reaches as stationary distribution of states exactly the posterior $p(\omega|X,Y)$ that we need, by only knowing that $p(\omega|X,Y)\propto p(Y|X,\omega)p(\omega)$.

The first and simplest algorithm to achieve this is the Metropolis-Hasting algorithm~\cite{metropolis1953equation}, but in the present work we will use the No-U-Turn Sampler (NUTS)~\cite{hoffman2014no}, which is a variant of the Hamiltonian Monte Carlo (HMC)~\cite{duane1987hybrid}.

In HMC the sampling process is represented by the dynamics of a Hamiltonian system, where:
\begin{itemize}
    \item the position $q$ is the state of the Markov chain itself, i.e., $\omega$,
    \item an artificial momentum variable $p$ is introduced,
    \item the Hamiltonian of the system is $H(q,p)=K(p)+U(q)$, where a possible choice for the kinetic energy is $K(p)=p^TM^{-1}p/2$ (with $M$ a symmetric, positive-definite \enquote{mass}), and the potential energy is given by
        \begin{equation*}
            U(q)=-\log p(\omega|X,Y) = -\log p(Y|X,\omega)p(\omega) + \text{const},
        \end{equation*}
        where the integral~\eqref{eq:normalisation} can be removed because it will not affect the dynamics of the fictitious system above.
\end{itemize}
The equations of motion are then
\begin{align*}
    \frac{d q_i}{dt} &= \frac{\partial H}{\partial p_i} = [M^{-1}p]_i \\
    \frac{d p_i}{dt} &= -\frac{\partial H}{\partial q_i} = -\frac{\partial U}{\partial q_i},
\end{align*}
which can be numerically solved with the leapfrog method.
The algorithm proceeds to sample as follows:
\begin{enumerate}
    \item a momentum $p$ is sampled from its Gaussian distribution $f(p)=e^{K(p)/T}/Z$, where $T$ is a temperature, $Z$ is the normalisation and $M$ is the covariance matrix;
    \item a new state is proposed using Hamiltonian dynamics: from current state $(q,p)$ we take $L$ steps with the leapfrog method with step-size $\varepsilon$ and we get to $(q^*,p^*)$; 
    \item the new state is accepted or rejected according to $\min[1,\exp(-H(q^*,p^*)+H(q,p))]$.
\end{enumerate}
One can prove that HMC is ergodic and the stationary distribution it converges to is exactly $p(\omega|X,Y)$.

\bibliographystyle{abbrv}


\bibliography{sample}

\end{document}